\documentclass[11pt]{article}

\usepackage[final]{acl}

\usepackage{times}
\usepackage{latexsym}
\usepackage{amsmath}
\usepackage{threeparttablex}
\usepackage[T1]{fontenc}

\usepackage[utf8]{inputenc}

\usepackage{microtype}

\usepackage{inconsolata}

\usepackage{graphicx}

\usepackage{multirow}
\usepackage{booktabs}
\usepackage{cuted}
\usepackage{caption}

\usepackage{tikz}
\usetikzlibrary{shapes.geometric, arrows.meta, positioning, fit, backgrounds, calc}

\definecolor{inputblue}{HTML}{DBEAFE}
\definecolor{inputborder}{HTML}{3B82F6}
\definecolor{llmyellow}{HTML}{FEF3C7}
\definecolor{llmborder}{HTML}{F59E0B}
\definecolor{toolgreen}{HTML}{D1FAE5}
\definecolor{toolborder}{HTML}{10B981}
\definecolor{validred}{HTML}{FEE2E2}
\definecolor{validborder}{HTML}{EF4444}
\definecolor{statepurple}{HTML}{EDE9FE}
\definecolor{statebordr}{HTML}{8B5CF6}
\definecolor{loopbg}{HTML}{F8FAFC}
\definecolor{loopborder}{HTML}{94A3B8}
\definecolor{feedbackred}{HTML}{DC2626}

\definecolor{inputgray}{HTML}{F3F4F6}
\definecolor{inputgrayborder}{HTML}{9CA3AF}
\definecolor{agentorange}{HTML}{FFF7ED}
\definecolor{agentborder}{HTML}{F97316}
\definecolor{fig2toolblue}{HTML}{DBEAFE}
\definecolor{fig2toolborder}{HTML}{3B82F6}
\definecolor{validpurple}{HTML}{F3E8FF}
\definecolor{decisionyellow}{HTML}{FEF9C3}
\definecolor{decisionborder}{HTML}{EAB308}
\definecolor{shortpink}{HTML}{FFE4E6}
\definecolor{shortborder}{HTML}{F43F5E}
\definecolor{stategreen}{HTML}{DCFCE7}
\definecolor{fig2stateborder}{HTML}{22C55E}
\definecolor{futuregray}{HTML}{E0F2FE}
\definecolor{futureborder}{HTML}{38BDF8}
\definecolor{loopcolor}{HTML}{F97316}
\definecolor{contextcolor}{HTML}{22C55E}
\definecolor{stateloopcolor}{HTML}{6B7280}

\title{ReacTOD: Bounded Neuro-Symbolic Agentic NLU for Zero-Shot Dialogue State Tracking}

\author{
  \textbf{Yanjun Lin\textsuperscript{*}} \quad
  \textbf{Zimo Xiao\textsuperscript{*}} \quad
  \textbf{Kartik Natarajan} \quad
  \textbf{Mahesh Sankaranarayanan} \quad
  \textbf{Niraj Nawanit} \\
  \textbf{Rakshit Parashar} \quad
  \textbf{Austin Zhang} \quad
  \textbf{Karthik Konaraddi} \quad
  \textbf{Rishita Mote} \quad
  \textbf{Wei Niu} \\
  \\
  Amazon \\
  \texttt{\{linyj, zimoxiao, kartikn, sankmahe, nawanit,} \\
  \texttt{chillorb, auszhang, kartkon, 
rmote, niuwei\}@amazon.com}
}

\begin{document}
\maketitle
\renewcommand{\thefootnote}{\fnsymbol{footnote}}
\footnotetext[1]{Equal contribution.}
\renewcommand{\thefootnote}{\arabic{footnote}}
\begin{strip}
  \vspace{5pt}
  \centering
  \begin{tikzpicture}[
    scale=0.78, transform shape,
    node distance=0.8cm and 1.2cm,
    >={Stealth[length=3mm]},
    every node/.style={font=\small},
    box/.style={draw, rounded corners=4pt, minimum height=1.1cm, minimum width=2.4cm, align=center, line width=0.8pt},
  ]
  \node[box, fill=inputblue, draw=inputborder] (input) {
      \textbf{Dialogue Input}\\[-1pt]
      {\scriptsize $u_t$, $a_{t-1}$, $B_{t-1}$}
  };
  \node[box, fill=llmyellow, draw=llmborder, right=1.8cm of input, minimum width=3cm] (llm) {
      \textbf{LLM Engine}\\[-1pt]
      {\scriptsize 1. Reason (Thought)}\\[-2pt]
      {\scriptsize 2. Select Tool ($a_k$)}
  };
  \node[box, fill=validred, draw=validborder, right=1.5cm of llm, minimum width=3cm] (valid) {
      \textbf{Deterministic Validator}\\[-1pt]
      {\scriptsize 1. Action compliance}\\[-2pt]
      {\scriptsize 2. Schema conformance}\\[-2pt]
      {\scriptsize 3. Coreference consistency}
  };
  \node[box, fill=toolgreen, draw=toolborder, right=1.5cm of valid, minimum width=2.8cm] (tools) {
      \textbf{Tool Library $\mathcal{T}$}\\[-1pt]
      {\scriptsize $\tau_{IC}$: Intent Classification}\\[-2pt]
      {\scriptsize $\tau_{SR}$: Slot Resolution}\\[-2pt]
      {\scriptsize $\tau_{H}$: History Retrieval}
  };
  \node[box, fill=statepurple, draw=statebordr, right=1.5cm of tools, minimum width=2.4cm] (state) {
      \textbf{Belief State $B_t$}\\[-1pt]
      {\scriptsize Deferred upsert-only}\\[-2pt]
      {\scriptsize update}
  };
  \draw[->, line width=0.8pt] (input) -- (llm);
  \draw[->, line width=0.8pt] (llm) -- node[above, font=\scriptsize] {tool call} (valid);
  \draw[->, line width=0.8pt, color=toolborder] (valid) -- node[above, font=\scriptsize, color=toolborder] {pass} (tools);
  \draw[->, line width=0.8pt, color=statebordr] (tools) -- node[above, font=\scriptsize, color=statebordr] {$\tau_{SR}$ result} (state);
  \draw[->, line width=0.8pt, color=feedbackred, dashed]
      (valid.south) -- ++(0, -0.5cm) -| node[below, pos=0.3, font=\scriptsize, color=feedbackred, align=center] {error feedback} (llm.south);
  \draw[->, line width=0.8pt, color=loopborder]
      (tools.north) -- ++(0, 0.6cm) -| node[below, pos=0.25, font=\scriptsize, color=gray!60!black] {tool result $\rightarrow$ next iteration} (llm.north);
  \begin{scope}[on background layer]
  \node[draw=loopborder, dashed, line width=1pt, rounded corners=8pt,
        fill=loopbg, fill opacity=0.3,
        fit=(llm)(tools)(valid),
        inner sep=26pt,
        inner xsep=20pt,
        yshift=0.1cm,
        label={[font=\small\bfseries, anchor=north west, yshift=-2pt, xshift=5pt]north west:{Bounded ReAct Loop ($\leq K_{max}$ iterations)}}] (loop) {};
  \end{scope}
  \node[below=0.4cm of input, font=\scriptsize\itshape, color=gray!60!black, align=center] {
      Dynamic context:\\[-1pt]
      $B_{t-1}$ + $a_{t-1}$ + $u_t$
  };
  \node[below=0.4cm of state, font=\scriptsize\itshape, color=gray!60!black, align=center] {
      Monotonic accumulation\\[-1pt]
      (multi-domain)
  };
  \end{tikzpicture}
  \vspace{-1.0em}
  
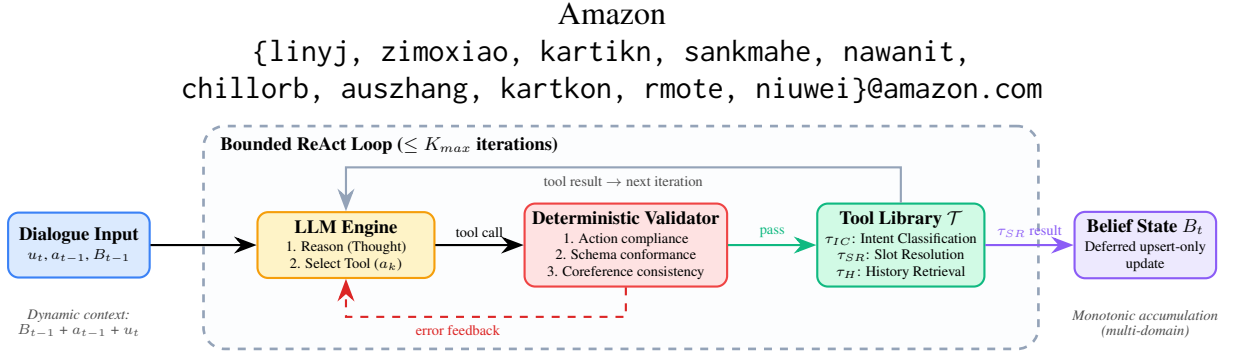
\captionof{figure}{Overview of the ReacTOD bounded neuro-symbolic agentic architecture.}
  \label{fig:overview}
  \vspace{0.5em}
\end{strip}

\begin{abstract}

Task-oriented dialogue systems---handling transactions, reservations, and service requests---require predictable behavior, yet the moderately-sized LLMs needed for practical latency are prone to hallucination and format errors that cascade into incorrect actions (e.g., a hotel booked for the wrong date). We propose \textbf{ReacTOD}, a bounded neuro-symbolic architecture that reformulates NLU as discrete tool calls within a self-correcting ReAct loop governed by deterministic validation. A bounded ReAct loop enables iterative self-correction, improving accuracy by up to 9.3 percentage points over single-pass inference on MultiWOZ. A symbolic validator enforces action compliance, schema conformance, and coreference consistency on every dialogue state update, achieving a 93.1\% self-correction rate on intercepted errors and producing structured execution traces. Incremental state prediction and on-demand history retrieval keep prompts compact, empirically improving instruction adherence in parameter-constrained models. On MultiWOZ 2.1, ReacTOD achieves a new zero-shot state-of-the-art: gpt-oss-20B reaches 52.71\% joint goal accuracy, surpassing the previous best by 14 percentage points, while Qwen3-8B achieves 47.34\% with only 8B parameters. On the Schema-Guided Dialogue (SGD) benchmark, ReacTOD with Claude-Opus-4.6 achieves 80.68\% JGA under fully end-to-end evaluation with predicted domains, and Qwen3-32B reaches 64.09\%---demonstrating cross-benchmark generalization without task-specific training data.
\end{abstract}

\section{Introduction}
Task-Oriented Dialogue (TOD) systems deployed in production environments---handling hotel bookings, restaurant reservations, and transport arrangements---require predictable, verifiable NLU behavior: an incorrectly resolved slot value (e.g., a check-in date inferred from the wrong turn) propagates to downstream API calls, producing silent failures or incorrect transactions. This need for reliable outputs has historically driven the dominance of discriminative, pipelined Natural Language Understanding (NLU) architectures, where extractive models like BERT perform Intent Classification (IC) and Slot Resolution (SR) as sequential tasks over fixed label sets. While these pipelines offer sub-second latency and high predictability, they depend on large volumes of domain-specific labeled data and require retraining to accommodate new intents or language variation---limiting zero-shot generalization.

To move beyond static ontologies, recent work has shifted toward generative, LLM-driven prompting for zero-shot NLU. Frameworks such as FnCTOD~\citep{li2024fnctod} reframe domain logic as executable functions, leveraging in-context learning. However, single-pass generative approaches suffer from probabilistic variance and faithful hallucinations, in which the model confidently infers unstated entity values to complete a schema, posing significant risks in production dialogue pipelines where incorrect state values propagate to downstream API calls. Moreover, the linguistic complexities of real dialogue (cross-turn coreference, implicit value acceptance, non-linear domain switching) require multi-step reasoning and dynamic context retrieval. Unbounded agentic frameworks can address these phenomena in principle, but their reliance on open-ended reasoning loops and frontier-scale models introduces impractical latency and computational overhead.

We argue that improving the reliability of LLM-based DST does not primarily require larger models, but rather stronger architectural control over the reasoning process. Our key insight is that LLM errors in dialogue state tracking are predominantly local and correctable---a misformatted time value or an invalid slot name, rather than a fundamental misunderstanding of the dialogue. Based on this insight, we propose \textbf{ReacTOD}, a hybrid neuro-symbolic NLU architecture that decomposes NLU into discrete tool calls within a bounded ReAct-style reasoning loop, reducing the per-step burden on the LLM. A deterministic validator intercepts all tool outputs before any state mutation, enforcing action compliance, schema conformance, and coreference consistency---enabling the model to self-correct from structured error feedback rather than requiring re-processing of the entire dialogue. This constrained design lowers the reasoning capacity required per step, enabling parameter-efficient models (e.g., Qwen3-8B) to achieve robust agentic state tracking without frontier-scale compute. The architectural details are presented in Section~\ref{sec:methodology} and illustrated in Figure~\ref{fig:overview}.

We evaluate our architecture on MultiWOZ 2.1 and the Schema-Guided Dialogue (SGD) benchmark in a zero-shot setting---no labeled dialogues, no fine-tuning, no in-domain examples---using dynamic schema injection across five backbone models ranging from 8B to frontier scale. On MultiWOZ 2.1, ReacTOD with gpt-oss-20B achieves 52.71\% Joint Goal Accuracy (JGA), surpassing the previous zero-shot state-of-the-art (FnCTOD with GPT-4, 38.71\%) by 14 percentage points (pp). Even Qwen3-8B reaches 47.34\% JGA---exceeding FnCTOD with the 4$\times$ larger Qwen3-32B (40.36\%)---demonstrating that the gains stem from architectural design rather than model scale. On SGD, ReacTOD with Claude-Opus-4.6 achieves 80.68\% JGA, outperforming a reproduced SRP baseline (45.20\%) that uses gold domain labels, and the ReAct loop contributes up to 11.82 pp over single-pass inference, confirming cross-benchmark generalization without task-specific training data.
In summary, our contributions are threefold:
\begin{enumerate}
\setlength{\itemsep}{0pt}
\setlength{\parskip}{0pt}
\item \textbf{Bounded Agentic Reasoning:} We introduce a constrained ReAct architecture that decomposes NLU into discrete tool calls with iterative self-correction, enabling error recovery beyond what single-pass inference achieves---with ablations showing gains of up to 9.3 percentage points.
\item \textbf{Deterministic Validation:} We design a symbolic validator that gates all state mutations, enforcing action compliance, schema conformance, and coreference consistency to catch common LLM errors (e.g., invalid tool calls, hallucinated slot names) before they reach the dialogue state.
\item \textbf{Parameter-Efficient Zero-Shot DST:} We demonstrate that incremental state prediction and on-demand context retrieval keep prompts compact, enabling models as small as 8B parameters to surpass prior zero-shot baselines built on larger LLMs. ReacTOD establishes a new state-of-the-art on MultiWOZ 2.1 and strong cross-benchmark performance on SGD without task-specific training data, requiring only a machine-readable domain schema.
\end{enumerate}

\section{Related Work}

\subsection{From Pipelined NLU to Generative State Tracking}

Early enterprise NLU treated IC and SR as sequential classification tasks over fixed label sets. JointBERT~\citep{chen2019bert} unified both through a shared encoder, while lightweight variants targeted resource-constrained devices~\citep{huang2022fast}. These discriminative pipelines offer determinism but are tied to predefined vocabularies, degrading on out-of-distribution inputs and precluding zero-shot generalization. Generative sequence-to-sequence models relaxed this constraint: TRADE~\citep{wu2019transferable} enabled cross-domain slot transfer via pointer-generator networks, and SimpleTOD~\citep{hosseini2020simple} and SOLOIST~\citep{peng2020soloist} consolidated the pipeline into a single autoregressive objective. However, these approaches still required substantial in-domain fine-tuning, leaving zero-shot adaptability an open challenge.

\subsection{LLM-Driven Prompting and Knowledge Distillation}

Instruction-tuned LLMs enabled zero-shot DST via in-context learning. D3ST~\citep{zhao2022description} and \citet{lu2024prompt} replaced schema notations with natural language descriptions, SERI-DST~\citep{lee2024seridst} dynamically retrieved dialogue examples, and FnCTOD~\citep{li2024fnctod} established the zero-shot state-of-the-art by treating domains as executable functions. However, single-pass generative approaches suffer from probabilistic variance and faithful hallucinations---confidently inferring unstated entity values to complete a schema~\citep{ji2023survey}---posing reliability risks in production dialogue systems where incorrect state values propagate to downstream API calls. Knowledge distillation approaches~\citep{xu2025schema,aguirre2024finetuning} reduce inference costs by training smaller student models on LLM-generated data, but hard-code the schema into model weights, sacrificing zero-shot flexibility.

\subsection{Tool-Augmented Agents and Neuro-Symbolic Integration}

ReAct~\citep{yao2023react} demonstrated that LLMs can interleave reasoning traces with task-specific actions, but deploying unbounded agents in TOD introduces reliability risks. \citet{elizabeth2025exploring} showed that ReAct-based agents frequently underperform structured baselines on task success metrics despite producing fluent responses, and while LLMs exhibit self-refinement capacity~\citep{madaan2023selfrefine}, unconstrained self-evaluation is susceptible to confirmation bias without external grounding. These findings motivate our core design principle: confining the LLM to narrowly scoped tool-mediated subtasks and gating all state mutations through a deterministic symbolic validator.

\section{Methodology}
\label{sec:methodology}

Our key insight is that LLM errors in dialogue state tracking are predominantly local and correctable---a misformatted time value or an invalid slot name, rather than a fundamental misunderstanding of the dialogue. By decomposing NLU into discrete tool calls within a bounded ReAct loop, we enable the agent to receive structured feedback from a deterministic validator and repair such mistakes iteratively, without requiring the model to re-process the entire dialogue context. This principle motivates a bounded neuro-symbolic architecture that separates the Natural Language Understanding (NLU) pipeline into isolated, verifiable tasks, orchestrates them via a constrained ReAct-style state machine, and gates all state mutations through deterministic validation.

\subsection{Problem Formulation and Architecture Overview}

\begin{figure*}[t]
  \centering
  \begin{tikzpicture}[
    scale=0.82, transform shape,
    node distance=0.4cm and 1.2cm,
    >={Stealth[length=2.5mm]},
    every node/.style={font=\small},
    ctxbox/.style={draw, rounded corners=3pt, minimum width=3.2cm, align=left, line width=0.7pt, inner sep=5pt},
    agentbox/.style={draw, rounded corners=3pt, minimum width=2.8cm, minimum height=2cm, align=center, line width=0.7pt, inner sep=5pt},
    toolbox/.style={draw, rounded corners=3pt, minimum width=3.8cm, align=left, line width=0.7pt, inner sep=4pt},
    valbox/.style={draw, rounded corners=3pt, minimum width=2.8cm, align=center, line width=0.7pt, inner sep=5pt},
    statebox/.style={draw, rounded corners=3pt, minimum width=2.4cm, align=center, line width=0.7pt, inner sep=5pt},
  ]

  \node[ctxbox, fill=inputgray, draw=inputgrayborder] (sysprompt) {
      \textbf{System Prompt}\\[2pt]
      {\scriptsize \textbullet\ Instructions}\\[-1pt]
      {\scriptsize \textbullet\ Intent list}\\[-1pt]
      {\scriptsize \textbullet\ Tool definitions}
  };

  \node[ctxbox, fill=inputblue, draw=inputborder, below=0.3cm of sysprompt] (context) {
      \textbf{Dialogue Context}\\[2pt]
      {\scriptsize \textbullet\ $u_t$ (utterance)}\\[-1pt]
      {\scriptsize \textbullet\ $a_{t-1}$ (sys action)}\\[-1pt]
      {\scriptsize \textbullet\ $B_{t-1}$ (belief state)}\\[-1pt]
      {\scriptsize \textbullet\ $i_{t-1}$ (active intent)}
  };

  \node[agentbox, fill=llmyellow, draw=llmborder, right=1.5cm of sysprompt, yshift=-1cm] (llm) {
      \textbf{LLM Agent}\\[2pt]
      {\scriptsize ReAct loop:}\\[-1pt]
      {\scriptsize Thought $\rightarrow$}\\[-1pt]
      {\scriptsize Action $\rightarrow$}\\[-1pt]
      {\scriptsize Observation}
  };

  \node[valbox, fill=validred, draw=validborder, right=1.2cm of llm] (validator) {
      \textbf{Validator}\\[2pt]
      {\scriptsize Action}\\[-1pt]
      {\scriptsize Schema}\\[-1pt]
      {\scriptsize Coreference}
  };

  \node[toolbox, fill=toolgreen, draw=toolborder, right=1.2cm of validator, yshift=1.2cm] (toolIC) {
      \textbf{$\tau_{IC}$: Intent Classification}\\[1pt]
      {\scriptsize \textit{In:} intent $i_t$ }\\[-1pt]
      {\scriptsize \textit{Out:}  slot defs of $i_t$}
  };

  \node[toolbox, fill=toolgreen, draw=toolborder, below=0.3cm of toolIC] (toolSR) {
      \textbf{$\tau_{SR}$: Slot Resolution}\\[1pt]
      {\scriptsize \textit{In:} $\{(v_{raw}, v_{norm})\}$}\\[-1pt]
      {\scriptsize \textit{Out:} state update $\Delta B_t$}
  };

  \node[toolbox, fill=toolgreen, draw=toolborder, below=0.3cm of toolSR] (toolH) {
      \textbf{$\tau_{H}$: History Retrieval}\\[1pt]
      {\scriptsize \textit{In:} $n$ (turns)}\\[-1pt]
      {\scriptsize \textit{Out:} last $n$ turns of conversation}
  };

  \node[statebox, fill=statepurple, draw=statebordr, right=1cm of toolSR] (state) {
      \textbf{State $B_t$}\\[1pt]
      {\scriptsize Deferred}\\[-1pt]
      {\scriptsize upsert-only}
  };


  \draw[-, line width=0.7pt] (sysprompt.east) -- ++(0.4cm,0) -- ++(0,-1cm) coordinate (merge);
  \draw[-, line width=0.7pt] (context.east) -- ++(0.4cm,0) |- (merge);
  \draw[->, line width=0.7pt] (merge) -- (llm.west);

  \draw[->, line width=0.7pt, color=agentborder] (llm.east) -- node[above, font=\scriptsize, color=agentborder] {tool call} (validator.west);

  \draw[->, line width=0.7pt, color=toolborder] (validator.east) -- ++(0.4cm,0) |- (toolIC.west);
  \draw[->, line width=0.7pt, color=toolborder] (validator.east) -- ++(0.4cm,0) |- (toolSR.west);
  \draw[->, line width=0.7pt, color=toolborder] (validator.east) -- ++(0.4cm,0) |- (toolH.west);
  \node[above right=-0.2cm and 0.32cm of validator.east, font=\scriptsize, color=toolborder] {pass};

  \draw[->, line width=0.6pt, color=shortborder, dashed]
      (validator.south) -- ++(0,-1.5cm) -| node[below, pos=0.25, font=\scriptsize\itshape, color=shortborder] {error feedback} (llm.south);

  \draw[->, line width=0.6pt, color=toolborder, dashed]
      (toolIC.north) -- ++(0,0.5cm) -| node[above, pos=0.25, font=\scriptsize, color=toolborder] {slot defs} ([xshift=0.2cm]llm.north);

  \draw[->, line width=0.6pt, color=toolborder, dashed]
      (toolH.south) -- ++(0,-0.4cm) -| node[below, pos=0.25, font=\scriptsize, color=toolborder] {history} ([xshift=-0.2cm]llm.south);

  \draw[->, line width=0.7pt, color=statebordr] (toolSR.east) -- node[above, font=\scriptsize, color=statebordr] {update} (state.west);

  \end{tikzpicture}
  \caption{Tool definitions and data flow in ReacTOD. The validator checks all tool calls before execution; $\tau_{IC}$ and $\tau_H$ results feed back to the agent for the next step; only validated $\tau_{SR}$ output updates the belief state $B_t$. Invalid calls trigger error feedback to the LLM for self-correction.}
  \label{fig:control-flow}
\end{figure*}
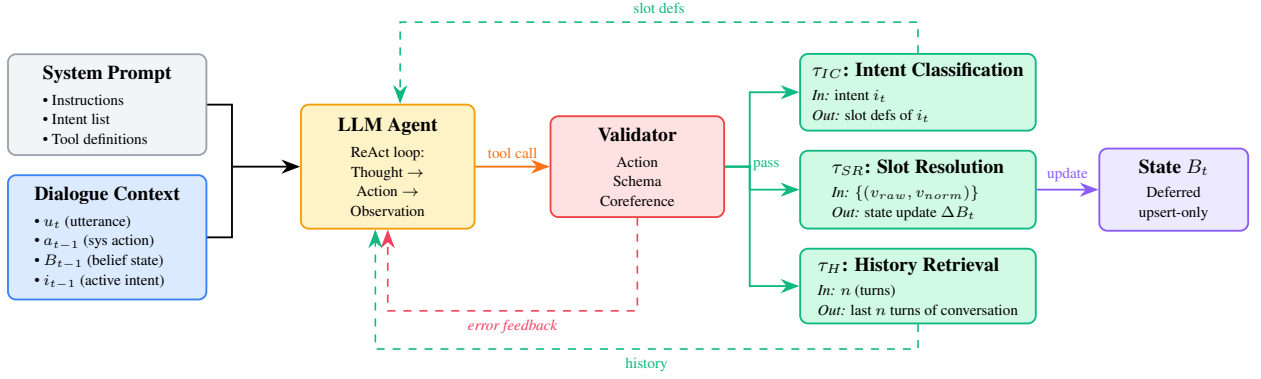

Given a dialogue turn consisting of user utterance $u_t$, prior system action $a_{t-1}$, persistent belief state $B_{t-1}$, and intents $i_{t-1}$, our architecture formulates NLU as a sequential, tool-augmented policy $\pi(a \mid s)$ whose action space $\mathcal{A}$ is restricted to tool library $\mathcal{T}$:
\begin{equation}
  a_k \sim \pi(\cdot \mid u_t, a_{t-1}, B_{t-1}, i_{t-1}, H_{<k}), \quad a_k \in \mathcal{T}
\end{equation}

\noindent where $H_{<k}$ denotes the agent's action--observation trace up to reasoning step $k$. Unlike single-pass approaches that jointly predict intent, slots, and schema formatting, this decomposition isolates each sub-task into a separate, verifiable tool invocation. As illustrated in Figure~\ref{fig:control-flow}, tool calls are first validated by a deterministic validator $V$ (\S\ref{sec:gatekeeper}) before execution, and only validated $\tau_{SR}$ results are permitted to update $B_t$, transforming state tracking from unbounded sequence generation into a bounded neuro-symbolic verification loop.

\textbf{Incremental Belief State Prediction.} To reduce per-turn complexity, the model predicts only incremental updates $\Delta B_t$ (newly mentioned or changed slots), with the full state recovered as $B_t = B_{t-1} \cup_{upsert} \Delta B_t$.

\textbf{Dynamic Context Construction.} Prior work has shown that shorter, focused prompts improve instruction adherence in smaller LLMs~\citep{xu2025evaluating}. Motivated by this finding, we construct each tool's context window dynamically rather than providing the full schema and dialogue history upfront. Intent definitions are included in the system prompt, but slot descriptions are injected only for the active intent at the time of slot resolution, avoiding irrelevant schema noise. Conversation history is not included by default; instead, the agent retrieves it on demand via a dedicated history tool ($\tau_H$) only when coreference resolution requires prior context. This lazy loading strategy keeps prompts minimal---typically containing only the active schema, current belief state $B_{t-1}$, previous system utterance $a_{t-1}$, and current user utterance $u_t$---reducing cognitive load for parameter-efficient models.

\subsection{Neuro-Symbolic Task Subdivision}

Following the empirical finding of FnCTOD \citep{li2024fnctod} that decomposing NLU into separate function calls improves LLM accuracy over joint prediction, we functionally separate Intent Classification (IC) and Slot Resolution (SR) into distinct generative invocations (see Figure~\ref{fig:control-flow}). This subdivision constrains the LLM's search space per inference step, reducing cognitive load and allowing parameter-efficient models to achieve reasoning parity with monolithic frontier models.

\subsubsection{Schema-Driven Intent Classification (IC)}

The IC module maps the user utterance $u_t$ to a target intent $i_t \in \mathcal{I}$, where $\mathcal{I}$ represents the dynamic ontology of the enterprise system. To achieve zero-shot generalization---requiring no labeled dialogues or fine-tuning---we include the domain schema in the LLM's context window at inference time, yielding the IC policy:

\begin{equation}
 i_t \sim \pi_{IC}(\cdot \mid u_t, B_{t-1}, a_{t-1})
\end{equation}
 To ensure deterministic out-of-domain (OOD) routing, $\mathcal{I}$ mandates the inclusion of non-transactional classes, such as $i_{fallback}$.

\textbf{The Short-Circuit Efficiency Logic:} If the IC policy predicts a non-transactional intent, the system executes a programmatic short-circuit, bypassing the SR module entirely. This prevents the computational inefficiency of forcing expensive entity extraction on conversational acknowledgments.

\subsubsection{Slot Resolution (SR)}

Conditioned on a transactional intent $i_t$, the SR module extracts the relevant entity set $E_t$. For each extracted slot, the model produces a tuple $e = \langle v_{raw}, v_{norm} \rangle$, where $v_{raw}$ is the surface form as it appears in the dialogue and $v_{norm}$ is the canonical, system-compliant normalization (e.g., mapping ``tmrw'' to an ISO-8601 date). To keep the generative task focused, only the slot definitions associated with the active intent $i_t$ are injected into the prompt, excluding irrelevant schema from other domains.

To handle \textbf{Implicit Acceptance}---where a user confirms a system proposal without restating the entity (e.g., the system asks ``How about the Hilton?'' and the user responds ``Yes'')---the previous system utterance $a_{t-1}$ is included in the SR context window, allowing the model to ground extractions in the system's prior turn.

\subsection{Bounded Agentic Control Flow}

The agent operates within a single bounded ReAct loop with a maximum of $K_{max}$ iterations. The system prompt instructs the agent to follow a prescribed tool-calling sequence: first invoke $\tau_{IC}$ to classify the user's intent, then invoke $\tau_{SR}$ to extract and resolve slots for the predicted intent. While this ordering is enforced via prompt instruction rather than programmatic constraint, the deterministic validator (\S\ref{sec:gatekeeper}) provides a hard guarantee: $\tau_{SR}$ cannot successfully terminate the loop unless the extracted slots pass all validation checks for the predicted intent. This creates an effective control flow where the agent self-organizes into an IC-first, SR-second pattern, with the validator acting as the structural enforcement mechanism.

The agent may re-invoke $\tau_{IC}$ if extracted slots are inconsistent with the predicted intent, re-invoke $\tau_{SR}$ to correct slot values after validation feedback, or invoke $\tau_{H}$ to retrieve conversation history for coreference resolution. If the iteration limit $K_{max}$ is reached without successful validation, the system forces graceful degradation by returning a fallback response.

\subsection{Deterministic Validation and Self-Correction}
\label{sec:gatekeeper}

A central design principle of our architecture is that the LLM proposes actions but never directly modifies system state. We introduce a Deterministic Validator $V$, a deterministic gatekeeper that evaluates any proposed tool call $a_k$ and the current state $s_k$.

\begin{equation}
V(a_k, s_k) \rightarrow \begin{cases}
 (True, \emptyset) & \text{if safe} \\
 (False, \varepsilon_{feedback}) & \text{if violated}
\end{cases}
\end{equation}

The validator operates via deterministic symbolic checks to avoid the infinite regress problem commonly associated with LLM-as-a-judge frameworks.

\textbf{The Symbolic Validator:} This layer executes computationally cheap, $O(1)$ algorithmic checks organized into three categories: \textbf{\textit{action compliance}} (rejecting calls to undefined tools, enforcing prerequisite ordering such as requiring $\tau_{IC}$ before $\tau_{SR}$, and suppressing duplicate tool calls), \textbf{\textit{schema conformance}} (validating intent and slot names against the domain ontology, and enforcing value constraints including regex matching for dates, times, and numbers against canonical formats and enumeration membership checks), and \textbf{\textit{coreference consistency}} (flagging generic references such as ``restaurant'' that indicate unresolved entities requiring history retrieval). When a check fails, the validator generates a structured error message describing the violation and injects it back into the agent's context, prompting the model to self-correct on the next iteration. For example, given the utterance ``I need to be there on time for my reservation,'' if the model extracts a non-temporal value for slot \texttt{taxi-arriveby}, the validator rejects the output with feedback such as ``invalid format for slot \texttt{taxi-arriveby}: expected HH:MM,'' steering the model to re-examine the utterance and extract the correct time reference. To prevent infinite looping, the system enforces a maximum iteration threshold $K_{max}$. If this limit is reached without successful validation, the system forces graceful degradation by returning a fallback response, safely terminating the loop without further LLM inference.

\textbf{Deferred State Updates.} The belief state $B_t$ is a persistent, multi-domain table that supports non-linear context switching across domains. Crucially, $B_t$ is never mutated during the agent's reasoning iterations---updates are deferred until the validator confirms the extracted slots pass all checks. This isolation ensures that rejected intermediate outputs from self-correction attempts cannot corrupt the persistent state, preserving a consistent view of the dialogue for subsequent turns. Updates follow a strict upsert-only protocol: new slot-value pairs are inserted, existing slots are overwritten on user revision (e.g., ``Actually, make it for 3 people''), but values are never deleted (explicit removal uses a designated null value). This monotonic accumulation ensures that cross-domain state is preserved even as the conversation switches between domains.

\section{Experiment}

\subsection{Evaluation Datasets}
To evaluate the DST performance of our proposed methodology, we use two widely adopted multi-domain task-oriented dialogue benchmarks.
\paragraph{MultiWOZ 2.1.} The Multi-Domain Wizard-of-Oz 2.1 dataset~\citep{budzianowski2018multiwoz,eric2020multiwoz} contains human-to-human task-oriented conversations spanning five domains with cross-domain coreference. We evaluate on the 1,000-dialogue test split, using version 2.1 for comparability with prior work despite known annotation issues in later versions~\citep{zang2020multiwoz,ye2022multiwoz}.

\paragraph{Schema-Guided Dialogue (SGD).} The SGD dataset~\citep{rastogi2020towards} spans 26 services across 16 domains. Its schema-driven design---where each service defines its own intents, required/optional slots, and result slots---closely mirrors real-world API-driven systems. We evaluate on the 4,201-dialogue test split.

\subsection{Metrics}

We adopt Joint Goal Accuracy (JGA) as our primary metric. For MultiWOZ 2.1, following the TRADE protocol~\citep{wu2019transferable}, we report \textit{overall JGA} (exact match across all active domains simultaneously) and \textit{domain-specific JGA} (per-domain exact match). For SGD, we follow the official evaluation protocol~\citep{rastogi2020towards} with per-service JGA averaged across services. Non-categorical slot values are compared using fuzzy token-sort matching.

\subsection{Experiment Setups}
\subsubsection{Model Configuration}

To ensure our methodology is deployable in real-world applications with practical latency and cost constraints, we primarily evaluate on open-source LLMs with fewer than 32 billion parameters. We additionally require models to possess sufficient reasoning capability to operate within the ReAct loop and make appropriate action decisions. Based on these criteria, we evaluate the following models: Qwen3-8B, Qwen3-32B \citep{qwen3technicalreport}, gpt-oss-20B \citep{openai2025gptoss120bgptoss20bmodel}, and Gemma3-12B \citep{gemmateam2025gemma3technicalreport}. To further assess the upper bound of our system's performance, we also include Claude-Opus-4.6 as a high-capacity reference model.

All models are evaluated with a temperature of 0.0 to encourage deterministic and reproducible outputs and share a uniform maximum ReAct iteration cap of $K_{\max} = 6$. Thinking mode is disabled for Qwen3 models and Gemma3-12B (text-based ReAct prompting); gpt-oss-20B uses native thinking with low effort; Claude-Opus-4.6 interleaves free-form reasoning with native tool calls. Qwen3-32B and Claude-Opus-4.6 are served via Amazon Bedrock; remaining models are hosted locally on A100 GPUs with vLLM. All five backbone models are evaluated on both MultiWOZ 2.1 and SGD.

\subsubsection{Schema Configuration}

\paragraph{MultiWOZ.} We source schema information from MultiWOZ 2.2, which provides formal intent definitions and slot metadata absent from 2.1. We merge intents within each domain into a single intent and augment each slot with a type annotation (\textit{Categorical}, \textit{Time}, \textit{Number}, or \textit{Freeform Text}) for the Deterministic Validator. Slots such as service name and food type are treated as freeform text to reflect real-world conditions where exhaustive enumeration is impractical.

\paragraph{SGD.} We derive the schema programmatically from the official test set schema definitions~\citep{rastogi2020towards}. We retain separate intents per service and promote result-only slots to sibling search intents when needed. Each slot is annotated with a role---\textit{Required} or \textit{Filter}---derived from the schema's \texttt{required\_slots}, \texttt{optional\_slots}, and default values. Purely informational slots (appearing only in \texttt{result\_slots}) are excluded from the model's slot list to reduce hallucination. Date and time slots are normalized to canonical formats; other values are treated as freeform text.

\begin{table*}[t]
  \centering
  \small
  \begin{threeparttable}
    \begin{tabular}{llccc}
      \toprule
      & & \multicolumn{2}{c}{\textbf{MultiWOZ 2.1}} & \textbf{SGD} \\
      \cmidrule(lr){3-4} \cmidrule(lr){5-5}
      \textbf{Approach} & \textbf{Model} & \textbf{Overall JGA} & \textbf{Domain Avg.\ JGA} & \textbf{Avg.\ Svc.\ JGA} \\
      \midrule
      SERI-DST & GPT-3.5 & N/A & 60.58\% & --- \\
      FnCTOD & GPT-4 & 38.71\% & 62.59\% & --- \\
      FnCTOD & Llama2-13B\tnote{*} & 37.67\% & 59.54\% & --- \\
      FnCTOD & Qwen3-32B\tnote{**} & 40.36\% & 63.10\% & --- \\
      FnCTOD & gpt-oss-20B\tnote{**} & 34.03\% & 58.56\% & --- \\
      DistDST & Llama-3.1-8B\tnote{*} & 45.20\% & --- & --- \\
      SRP\tnote{\dag} & Claude-Opus-4.6\tnote{\dag\dag} & --- & --- & 45.20\% \\
      \midrule
      ReacTOD & Qwen3-8B & 47.34\% & 68.11\% & 57.31\% \\
      ReacTOD & Qwen3-32B & 51.53\% & \textbf{71.83\%} & \textbf{64.09\%} \\
      ReacTOD & gpt-oss-20B & \textbf{52.71\%} & 71.77\% & 62.92\%\\
      ReacTOD & Gemma3-12B & 45.11\% & 66.35\% & 55.58\% \\
      \midrule
      ReacTOD & Claude-Opus-4.6\tnote{***} & 61.29\% & 78.34\% & 80.68\% \\
      \bottomrule
    \end{tabular}
    \begin{tablenotes}
      \scriptsize
      \item[*] Models are fine-tuned in the original work.
      \item[**] Re-evaluated with the public FnCTOD code using 5-shot prompts.
      \item[***] Reference only; impractical for production latency.
      \item[\dag] Uses gold domain labels and per-domain isolated sessions; ReacTOD predicts domains end-to-end.
      \item[\dag\dag] Reproduced with the SRP codebase and prompts; the original paper reports 88.70\% with GPT-4-Turbo, which we were unable to reproduce (see \S\ref{sec:baseline}).
    \end{tablenotes}
  \end{threeparttable}
  \caption{\label{tab:zero-shot-dst}
    Zero-shot DST on MultiWOZ 2.1 and SGD. Best results in \textbf{bold}; --- = not evaluated.}
\end{table*}

\subsection{Baseline}
\label{sec:baseline}
We compare our method \textbf{ReacTOD} against representative methods in the LLM-based dialogue state tracking paradigm. We include \textbf{SERI-DST} \citep{lee2024seridst}, which dynamically retrieves in-context dialogue examples to guide the LLM at inference time. Most critically, we compare against \textbf{FnCTOD} \citep{li2024fnctod}, which represents the previous zero-shot state-of-the-art by treating domain logic as executable functions and constraining the LLM to produce structured JSON arguments. To ensure a fair and up-to-date comparison, we additionally re-evaluate FnCTOD using its publicly available implementation with the same backbone LLMs used in our experiments (Qwen3-32B, gpt-oss-20B), isolating the contribution of our architectural design from differences in underlying model capability. We further include \textbf{DistDST}~\citep{xu2025schema}, a distillation-based method requiring offline fine-tuning (included for reference only). For SGD, we compare against \textbf{SRP}~\citep{safa2025zero}, which employs self-refined prompts with per-domain isolated chat sessions and gold domain labels. The original paper reports 88.70\% JGA with GPT-4-Turbo; our reproduction using the published SRP codebase with Claude-Opus-4.6 yields 45.20\% JGA after removing result-only slot predictions that the SRP prompt incorrectly extracts from system utterances. The dominant error is hallucination of informational attributes (e.g., \texttt{car\_name}, \texttt{venue}, \texttt{price}) caused by the prompt's instruction to track system-mentioned values. We report our reproduced result in Table~\ref{tab:zero-shot-dst}.

\section{Results}
We first report the overall zero-shot DST performance of \textbf{ReacTOD} against LLM-based baselines (\S\ref{sec:comparison}), then conduct an ablation study isolating the ReAct loop's contribution (\S\ref{sec:ablation}), followed by an efficiency and validator activation analysis (\S\ref{sec:latency}).

\subsection{Zero-Shot DST Performance}
\label{sec:comparison}

We first compare ReacTOD against baselines across models of varying capacity to assess whether the architectural gains hold independently of model scale. Table~\ref{tab:zero-shot-dst} presents the zero-shot DST results on MultiWOZ 2.1 and SGD. ReacTOD consistently outperforms all prompting baselines across both benchmarks. Even with Qwen3-8B, ReacTOD achieves 47.34\% Overall JGA and 68.11\% Domain Average JGA on MultiWOZ, surpassing FnCTOD with GPT-4 (38.71\% / 62.59\%) despite using a far smaller backbone. Our fair re-evaluation of FnCTOD with Qwen3-32B (40.36\%) further confirms that the gains stem from architectural design, not model capacity---ReacTOD with the smaller Qwen3-8B exceeds it by nearly 7 percentage points. The strongest MultiWOZ result comes from gpt-oss-20B at 52.71\% Overall JGA and 71.77\% Domain Average JGA, surpassing even the fine-tuned DistDST baseline (45.2\%). On SGD, which presents greater complexity (26 services across 16 domains, fine-grained schema distinctions), ReacTOD with Claude-Opus-4.6 achieves \textbf{80.68\%} average service JGA, outperforming the reproduced SRP baseline (45.20\%) despite operating under harder conditions---predicted domains and a single session, versus SRP's gold domain labels and per-domain isolation. All production-viable models (8B--32B) also surpass the SRP baseline.

Two cross-model comparisons reveal that reasoning capability matters more than parameter count within our framework. First, gpt-oss-20B matches the larger Qwen3-32B on MultiWOZ (52.71\% vs.\ 51.53\%) despite Qwen3-32B having 60\% more parameters, by leveraging native thinking mode and built-in tool calling for deeper per-step reasoning. On SGD, however, Qwen3-32B pulls ahead (64.09\% vs.\ 62.92\%), suggesting that the advantage of native tool calling diminishes when schema complexity increases and text-based reasoning suffices. Table~\ref{tab:efficiency} corroborates the cost of native thinking: gpt-oss-20B consumes 448 avg / 1611 P99 output tokens per turn versus 150 / 366 for Qwen3-32B. Second, Qwen3-8B overtakes the larger Gemma3-12B on both benchmarks (MultiWOZ: 47.34\% vs.\ 45.11\%; SGD: 57.31\% vs.\ 55.58\%) despite trailing without the ReAct loop (39.29\% vs.\ 42.73\%, Table~\ref{tab:ablation}). This crossover aligns with the Qwen3 technical report~\citep{qwen3technicalreport}, which shows Qwen3 outperforming Gemma-3 at comparable sizes on reasoning and agent benchmarks. These findings suggest that a model's capacity for structured multi-step reasoning is a stronger predictor of agentic DST performance than raw parameter count.

\begin{table*}[t]
  \centering
  \small
  \begin{tabular}{llccc}
    \toprule
    & & \multicolumn{2}{c}{\textbf{MultiWOZ 2.1}} & \textbf{SGD} \\
    \cmidrule(lr){3-4} \cmidrule(lr){5-5}
    \textbf{Model} & \textbf{Variant} & \textbf{Overall JGA} & \textbf{Domain Avg.\ JGA} & \textbf{Avg.\ Svc.\ JGA} \\
    \midrule
    \multirow{2}{*}{Qwen3-8B}
      & w/o ReAct Loop    & 39.29\% & 61.67\% & 45.49\% \\
      & ReacTOD           & 47.34\% {\small(+8.05)} & 68.11\% {\small(\textbf{+6.44})} & 57.31\% {\small(\textbf{+11.82})} \\
    \midrule
    \multirow{2}{*}{Qwen3-32B}
      & w/o ReAct Loop    & 46.35\% & 68.24\% & 56.36\% \\
      & ReacTOD           & 51.53\% {\small(+5.18)} & \textbf{71.83\%} {\small(+3.59)} & \textbf{64.09\%} {\small(+7.73)} \\
    \midrule
    \multirow{2}{*}{gpt-oss-20B}
      & w/o ReAct Loop    & 43.39\% & 66.44\% & 56.42\% \\
      & ReacTOD           & \textbf{52.71\%} {\small(\textbf{+9.32})} & 71.77\% {\small(+5.33)} & 62.92\%  {\small(+6.50)}\\
    \midrule
    \multirow{2}{*}{Gemma3-12B}
      & w/o ReAct Loop    & 42.73\% & 64.10\% & 53.34\% \\
      & ReacTOD           & 45.11\% {\small(+2.38)} & 66.35\% {\small(+2.25)} & 55.58\% {\small(+2.24)} \\
    \midrule
    \multirow{2}{*}{Claude-Opus-4.6}
      & w/o ReAct Loop    & 59.69\% & 77.89\% & 73.49\% \\
      & ReacTOD           & 61.29\% {\small(+1.60)} & 78.34\% {\small(+0.45)} & 80.68\% {\small(+7.19)} \\
    \bottomrule
  \end{tabular}
  \caption{Ablation study on MultiWOZ and SGD. ``w/o ReAct Loop'' = IC + SR as independent calls; (+) = gain in pp.}
  \label{tab:ablation}
\end{table*}

\subsection{Ablation Study}
\label{sec:ablation}

We next isolate \textit{bounded agentic reasoning} by ablating the bounded ReAct loop. We conduct experiments under two methodological conditions. In addition to the full \textbf{ReacTOD} pipeline, we evaluate a decomposed variant in which Intent Classification (IC) and Slot Resolution (SR) are issued as two independent LLM calls, without the iterative ReAct loop or thinking capability, serving as a direct ablation of the agentic reasoning component.

Table~\ref{tab:ablation} shows that the ReAct reasoning loop consistently yields substantial improvements across all backbone models. For Qwen3-8B, the full ReacTOD pipeline achieves 47.34\% Overall JGA and 68.11\% Domain Average JGA, compared to 39.29\% and 61.67\% for the decomposed variant---a gain of \textbf{8.05 pp} in Overall JGA. A similar pattern holds for gpt-oss-20B, where the full pipeline achieves 52.71\% / 71.77\% versus 43.39\% / 66.44\% without the ReAct loop, a gain of \textbf{9.32 pp}. The largest absolute improvement on MultiWOZ is observed with gpt-oss-20B, suggesting that models with native reasoning and tool-calling capabilities benefit most from the iterative self-correction mechanism. These results confirm that the iterative agentic reasoning loop is a critical architectural component of ReacTOD, enabling the model to self-correct slot resolution errors that single-pass inference cannot recover from.

The SGD ablation results confirm cross-benchmark generalization, with even larger gains than on MultiWOZ. Qwen3-8B gains 11.82 pp---the largest absolute improvement on either benchmark---suggesting that smaller models benefit disproportionately when the schema is more complex and the validator has more opportunities to catch errors. Gemma3-12B shows the smallest loop gain on both benchmarks, consistent with its weaker performance on reasoning and agentic benchmarks~\citep{gemmateam2025gemma3technicalreport,qwen3technicalreport}: the self-correction mechanism requires the model to interpret error feedback and revise its output, a capability that scales with reasoning proficiency.

\begin{table*}[t]
  \centering
  \small
  \begin{tabular}{lcccccc}
    \hline
    \multirow{2}{*}{\textbf{Model}}
      & \multicolumn{3}{c}{\textbf{LLM Calls / Turn}}
      & \multicolumn{3}{c}{\textbf{Output Tokens / Turn}} \\
    \cmidrule(lr){2-4} \cmidrule(lr){5-7}
    & \textbf{Avg} & \textbf{P50} & \textbf{P99} & \textbf{Avg} & \textbf{P50} & \textbf{P99} \\
    \hline
    Qwen3-8B        & 1.86 & 2.00 & 4.00  & 165.25  & 161.00 & 424.00 \\
    Qwen3-32B       & 1.83 & 2.00 & 4.00  & 150.40  & 151.00 & 365.58 \\
    gpt-oss-20B     & 1.85 & 2.00 & 3.00  & 448.09  & 386.00 & 1611.29 \\
    Gemma3-12B      & 2.19 & 2.00 & 6.00  & 161.69  & 147.00  & 557.90 \\
    Claude-Opus-4.6 & 1.89 & 2.00 & 3.00  & 207.45  & 199.00   & 629.58 \\
    \hline
  \end{tabular}
    \caption{\label{tab:efficiency}
    Efficiency of \textbf{ReacTOD} on MultiWOZ 2.1: LLM calls and output tokens per turn under full ReAct loop.
    }
\end{table*}

\subsection{Efficiency Analysis}
\label{sec:latency}

Finally, we analyze computational overhead to confirm the bounded loop remains practical. We report two hardware-agnostic proxy metrics---\textbf{LLM calls per turn} and \textbf{output tokens per turn}---rather than wall-clock latency, which varies with deployment configuration. Table~\ref{tab:efficiency} summarizes both metrics. Across all models, the median (P50) turn completes in exactly 2 LLM calls (the mandatory IC and SR invocations), with averages ranging from 1.83 to 2.19. P99 values of 3--6 calls confirm the loop remains bounded even in tail cases; Gemma3-12B exhibits the highest P99 (6.00), consistent with its weaker single-pass accuracy requiring more correction attempts.

Token consumption reflects differences in reasoning strategy: Qwen3 and Gemma3 models produce compact outputs (150--166 avg tokens) via text-based ReAct prompting, while gpt-oss-20B consumes roughly 3$\times$ more (448 avg, 1611 P99) due to native chain-of-thought tokens before each tool call. Claude-Opus-4.6 falls between these groups (207 avg), interleaving free-form reasoning with tool calls. The token overhead is thus determined by the backbone's reasoning strategy, not the agentic architecture.

\textbf{Validator Activation Analysis.} We next assess \textit{deterministic validation} by quantifying the validator's active role. We analyze Qwen3-8B, the smallest backbone where validation is most critical. Of 7,372 MultiWOZ turns, 683 (9.3\%) triggered at least one validator correction, producing 1,606 total feedback messages across three categories. \textit{Action compliance} dominates with 498 turns (6.8\%): the most frequent violation is attempting to submit slot values without first calling $\tau_{IC}$ (771 messages), followed by calls to undefined tools (222) and redundant duplicate invocations of the same tool within a turn (77). \textit{Schema conformance} accounts for 177 turns (2.4\%): invalid enumeration values such as ``Cambridge'' for an area slot constrained to \{centre, east, north, south, west\} (274 messages), hallucinated slot names like \texttt{attraction-postcode} or \texttt{train-departure-time} instead of the valid \texttt{train-leaveat} (58), and unrecognized intent names (47). \textit{Coreference consistency} triggers least frequently at 44 turns (0.6\%), flagging 157 generic entity references where the model outputs ``restaurant'' or ``hotel'' instead of the actual entity name, steering it to invoke $\tau_H$ for history retrieval. Of the 683 impacted turns, 636 self-corrected after structured feedback, with only 47 exhausting the $K_{max}=6$ ceiling---a \textbf{93.1\%} overall recovery rate (action 91.6\%, schema 91.5\%, coreference 95.5\%). To isolate the validator's contribution from the loop's iteration opportunity, we additionally evaluate a variant with the ReAct loop active but the validator disabled: Qwen3-8B drops from 47.34\% to 43.00\% JGA ($-$4.34 pp), confirming that structured error feedback---not merely additional inference attempts---drives the self-correction gains.

\section{Limitations}

While \textbf{ReacTOD} demonstrates strong zero-shot performance, several limitations warrant acknowledgment.

\paragraph{LLM Dependency and Cost.} ReacTOD introduces additional LLM calls per turn compared to single-pass approaches. While the ReAct loop is bounded in practice, each turn still incurs multiple inference requests, which may be a constraint in cost-sensitive or high-throughput production deployments.

\paragraph{Schema Dependency.} ReacTOD is zero-shot with respect to training data---no labeled dialogues, fine-tuning, or in-domain examples are required---but it does require a configured domain schema specifying intent definitions, slot names with descriptions, and type annotations. For schema-rich benchmarks like SGD, this configuration is largely derivable from existing service definitions. For MultiWOZ, where structured schema metadata is more limited, manual annotation of slot types was necessary. This schema engineering effort, while modest compared to collecting labeled training data, represents a setup cost that should be factored into deployment planning. Performance may further degrade in settings where the schema is incomplete, noisy, or absent, limiting applicability to truly open-domain dialogue.

\section{Conclusion}
We presented ReacTOD, a bounded neuro-symbolic architecture that decomposes NLU into discrete, validator-gated tool calls within a constrained ReAct loop. On MultiWOZ 2.1, ReacTOD establishes a new zero-shot state-of-the-art: gpt-oss-20B reaches 52.71\% Overall JGA, surpassing FnCTOD with GPT-4 (38.71\%) by 14 percentage points, while the 8B-parameter Qwen3-8B achieves 47.34\%---exceeding FnCTOD with the 4$\times$ larger Qwen3-32B. On SGD, ReacTOD with Claude-Opus-4.6 achieves 80.68\% per-service JGA under fully end-to-end evaluation (predicted domains), and Qwen3-32B reaches 64.09\%, confirming cross-benchmark generalization to SGD's 26-service, 16-domain schema space without task-specific training data. The agentic loop is the critical differentiator, contributing up to 9.3 pp on MultiWOZ and 11.82 pp on SGD over single-pass inference, yet remains computationally bounded: the median turn requires just two LLM calls with 150--207 avg output tokens. The deterministic symbolic validator catches and corrects common LLM errors---malformed values, invalid slot names, hallucinated entities---before they reach the dialogue state, achieving a 93.1\% self-correction rate on intercepted errors and producing a fully inspectable execution trace at every turn. By confining generative flexibility to narrowly scoped tasks and delegating control flow to symbolic logic, ReacTOD demonstrates that reliable zero-shot agentic NLU does not require frontier-scale models---structured reasoning capacity and deterministic safeguards matter more than parameter count.

\section*{Future Work}

\paragraph{Component-Isolation Ablations.} While we have isolated the validator's contribution via a controlled experiment (loop active, validator disabled; \S\ref{sec:latency}), controlled comparisons of lazy vs.\ full schema injection across model sizes---particularly on SGD where the 26-service schema amplifies context load---and always-on history inclusion vs.\ on-demand retrieval via $\tau_H$ remain as future work.

\paragraph{Extended Dialogue Management.} A natural extension of \textbf{ReacTOD} is to enrich the agent's tool repertoire beyond Intent Classification and Slot Resolution to support broader dialogue management capabilities. In particular, we plan to introduce tools for handling conversational control flows that arise frequently in real-world deployments: for instance, a \textit{wait} or \textit{clarification-pending} tool to gracefully manage turns where the system must defer resolution until additional user information is collected, and a \textit{repeat} or \textit{confirmation} tool to allow the agent to re-surface prior questions or confirm ambiguous slot values with the user. These additions would move \textbf{ReacTOD} closer to a fully agentic dialogue manager, capable of handling the full spectrum of conversational acts beyond state tracking.



\bibliography{custom}

\end{document}